\documentclass[journal=jcim,manuscript=article]{achemso}

\usepackage{times}
\usepackage{soul}
\usepackage{url}
\usepackage[hidelinks]{hyperref}
\usepackage[utf8]{inputenc}
\usepackage[T1]{fontenc}
\usepackage{graphicx}
\usepackage{amsmath,amssymb,amsfonts}
\usepackage{amsthm}
\usepackage{booktabs}
\usepackage{enumitem}
\usepackage[dvipsnames]{xcolor}
\usepackage{xspace}
\usepackage[flushleft]{threeparttable}
\usepackage{subcaption}
\usepackage{caption, mathtools}
\usepackage{comment}
\usepackage{algorithm}
\usepackage{algorithmic}

\SectionNumbersOn
\mciteErrorOnUnknownfalse

\newcommand{\method}{\mbox{$\mathop{\mathsf{RLSynC}}\limits$}\xspace}

\newcommand{\MDP}{\mbox{$\mathop{\mathsf{MDP}}\limits$}\xspace}

\newcommand{\augalg}{\mbox{$\mathsf{\method\text{-}aug}$}\xspace}
\newcommand{\topsearch}{\mbox{$\mathsf{\method\text{-}search}$}\xspace}

\newcommand{\reactant}{\mbox{$\mathop{\mathtt{R}}\limits$}\xspace}
\newcommand{\synthon}{\mbox{$\mathop{\mathtt{M}}\limits$}\xspace}
\newcommand{\product}{\mbox{$\mathop{\mathtt{P}}\limits$}\xspace}

\newcommand{\agent}{\mbox{$\mathop{\mathtt{A}}\limits$}\xspace}
\newcommand{\agentOne}{\mbox{$\mathop{\agent_1}\limits$}\xspace}
\newcommand{\agentTwo}{\mbox{$\mathop{\agent_2}\limits$}\xspace}

\newcommand{\action}{\mbox{$\mathop{\mathtt{a}}\limits$}\xspace}
\newcommand{\state}{\mbox{$\mathop{\mathtt{s}}\limits$}\xspace}
\newcommand{\statemb}{\mbox{$\mathop{\mathbf{h}}\limits$}\xspace}

\newcommand{\StateSpace}{\mbox{$\mathop{\mathcal{S}}\limits$}\xspace}
\newcommand{\ActionSpace}{\mbox{$\mathop{\mathcal{A}}\limits$}\xspace}
\newcommand{\Transition}{\mbox{$\mathop{\mathcal{T}}\limits$}\xspace}
\newcommand{\Reward}{\mbox{$\mathop{\mathcal{R}}\limits$}\xspace}

\newcommand{\ADD}{\mbox{$\mathop{\mathtt{ADD}}\limits$}\xspace}
\newcommand{\NOOP}{\mbox{$\mathop{\mathtt{NOOP}}\limits$}\xspace}

\newcommand{\retroformer}{\mbox{$\mathsf{RetroFormer}$}\xspace}
\newcommand{\retroprime}{\mbox{$\mathsf{RetroPrime}$}\xspace}

\newcommand{\gtworetro}{\mbox{$\mathsf{G^2Retro}$}\xspace}
\newcommand{\graphretro}{\mbox{$\mathsf{GraphRetro}$}\xspace}
\newcommand{\RPtoStoR}{\mbox{$\mathsf{RSMILES\text{-}p2s2r}$}\xspace}
\newcommand{\rsmiles}{\mbox{$\mathsf{RSMILES}$}\xspace}
\newcommand{\RPtoR}{\mbox{$\mathsf{RSMILES\text{-}p2r}$}\xspace}

\newcommand{\gtwogt}{\mbox{$\mathsf{G2GT}$}\xspace}
\newcommand{\mars}{\mbox{$\mathsf{MARS}$}\xspace}
\newcommand{\retroexplainer}{\mbox{$\mathsf{RetroExplainer}$}\xspace}
\newcommand{\grasp}{\mbox{$\mathsf{GRASP}$}\xspace}
\newcommand{\awac}{\mbox{$\mathsf{AWAC}$}\xspace}
\newcommand{\pex}{\mbox{$\mathsf{PEX}$}\xspace}
\newcommand{\rcsearcher}{\mbox{$\mathsf{RCSearcher}$}\xspace}
\newcommand{\gcpn}{\mbox{$\mathsf{GCPN}$}\xspace}
\newcommand{\moldqn}{\mbox{$\mathsf{MolDQN}$}\xspace}
\newcommand{\dynappo}{\mbox{$\mathsf{DynaPPO}$}\xspace}
\newcommand{\tcrppo}{\mbox{$\mathsf{TCRPPO}$}\xspace}

\newcommand{\map}{\mbox{{MAP@$N$}}\xspace}
\newcommand{\ndcg}{\mbox{{NDCG@$N$}}\xspace}

\newcommand{\diversity}{\mbox{{Diversity@$N$}}\xspace}

\newcommand{\data}{\mbox{$\mathcal{E}$}\xspace}
\newcommand{\validationset}{\mbox{$\mathcal{D}_{\text{validation}}$}\xspace}
\newcommand{\testset}{\mbox{$\mathcal{D}_{\text{test}}$}\xspace}
\newcommand{\trainset}{\mbox{$\mathcal{E}_{\text{train}}$}\xspace}
\newcommand{\trainaug}{\mbox{$\mathcal{E}_{\text{aug}}$}\xspace}
\newcommand{\randomset}{\mbox{$\mathcal{E}_{\text{random}}$}\xspace}

\newcommand{\reaction}{\mbox{$r$}\xspace}

\title{\method: Offline-Online Reinforcement Learning for Synthon Completion}

\author{Frazier N. Baker}
\affiliation{Computer Science and Engineering, The Ohio State University}
\author{Ziqi Chen}
\affiliation{Computer Science and Engineering, The Ohio State University}
\author{Daniel Adu-Ampratwum}
\affiliation{Division of Medicinal Chemistry and Pharmacognosy, The Ohio State University}
\author{Xia Ning}
\affiliation{Computer Science and Engineering, The Ohio State University}
\alsoaffiliation{Translational Data Analytics Institute, The Ohio State University}
\alsoaffiliation{Biomedical Informatics, The Ohio State University}
\email{ning.104@osu.edu}

\begin{document}

\maketitle

\begin{abstract}
Retrosynthesis is the process of determining the set of reactant molecules that can react to 
form a desired product.
Semi-template-based retrosynthesis methods, which imitate the reverse logic of synthesis reactions,
first predict the reaction centers in the products, and then complete the resulting synthons back
into reactants.
We develop a new offline-online reinforcement learning method \method for 
synthon completion in semi-template-based methods.
\method assigns one agent to each synthon, all of which complete the synthons by conducting actions 
step by step in a synchronized fashion. 
\method learns the policy from both offline training episodes and online interactions, which allows
\method to explore new reaction spaces.
\method uses a standalone forward synthesis model to evaluate the likelihood of the predicted reactants in 
synthesizing a product, and thus guides the action search.
Our results demonstrate that \method can outperform state-of-the-art synthon completion methods
with improvements as high as 14.9\%, highlighting its potential in synthesis planning.
\end{abstract}

\section{Introduction}
\label{sec:intro}

Retrosynthesis is the process of determining
the set of reactant molecules that can react to form a desired product molecule.
It is essential to drug discovery,
where medicinal chemists seek to identify feasible synthesis reactions for desired molecules 
(i.e., synthesis planning~\cite{chen_retrostar_2020}).
The recent development on computational retrosynthesis methods using deep learning~{\cite{seidl_improving_2022,tu_permutation_2022,zhong_root-aligned_2022,wan_retroformer_2022,yan2020retroxpert}} 
has enabled high-throughput and large-scale prediction for many products, facilitating medicinal chemists 
to conduct synthesis planning much more efficiently.  
Among the existing computational retrosynthesis methods, 
semi-template-based retrosynthesis methods~\cite{wang_retroprime_2021,somnath_learning_2021,chen_g2retro_2023} 
imitate the reverse logic of synthesis reactions: they first predict the reaction centers in the products, 
and then transform (complete) the resulting synthons -- 
the molecular structures from splitting the products at the reaction centers, 
back into reactants. 
Semi-template-based retrosynthesis methods enable necessary interpretability as to where the 
reactions happen among the reactants and how the products are synthesized, and thus have high practical 
utility to inform synthesis planning.

\begin{figure*}[!h]
\centering
\begin{minipage}[b]{.7\linewidth}
  \centering
	\includegraphics[width=\linewidth]{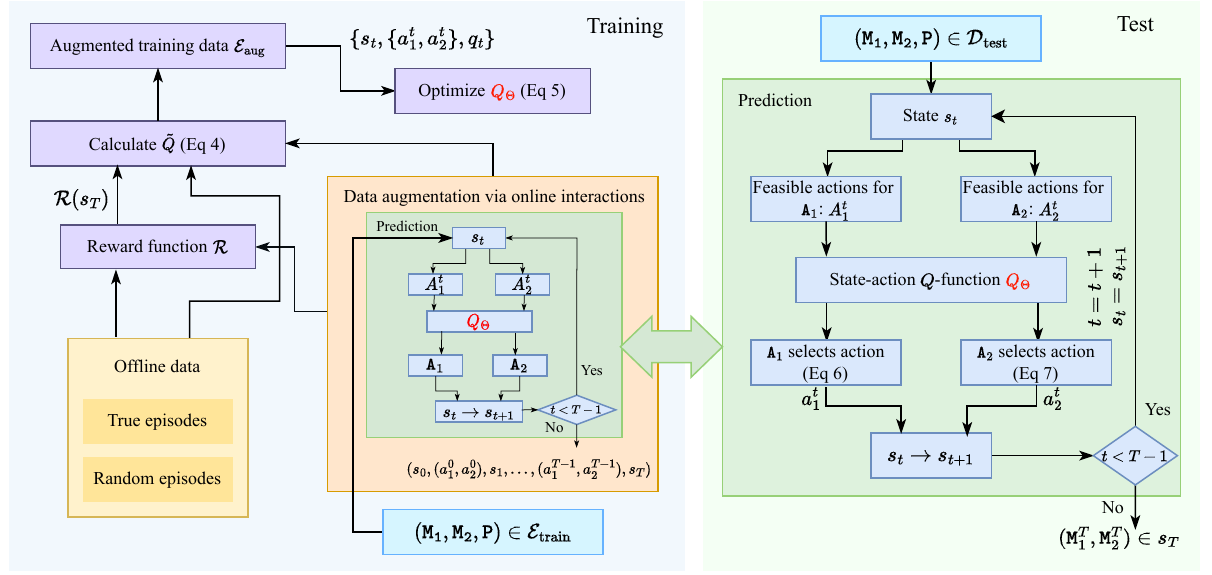}
		\caption{Overview Scheme of \method}
	\label{fig:overview}
\end{minipage}%
~
\begin{minipage}[b]{.3\linewidth}
  \centering
  \includegraphics[width=\linewidth]{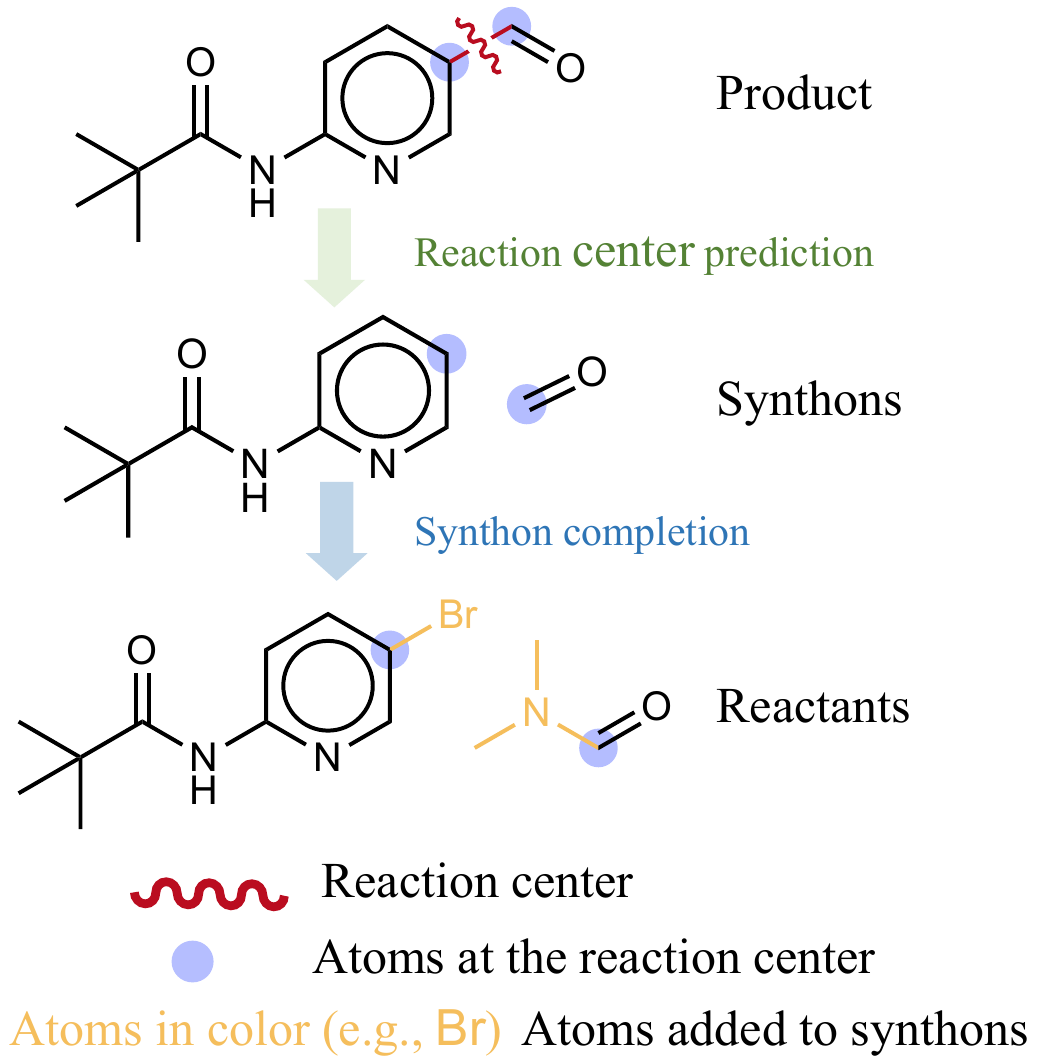}
	\caption{Retrosynthesis Process}
	\label{fig:retro}
\end{minipage}
\vspace{-20pt}
\end{figure*}

Existing retrosynthesis methods typically train predictive or generative models to transform from products 
to their reactants, under the supervision of training data with known reactions. 
The objective during model training is to reproduce the reactions in the training data, and the model performance 
is evaluated by comparing the predicted reactions of a product against its known reactions. 
While these models can accurately recover the known reactions for products, they do not have the capability of exploring and learning new reaction patterns not present in the training data. 
To address this issue, in this manuscript, we develop a new multi-agent reinforcement learning method 
with offline learning and online data augmentation, 
denoted as \method, for synthon completion in semi-template-based methods. 
Fig.~\ref{fig:overview} presents the overall idea of \method. 
We focus on semi-template-based methods due to their interpretability, practical utility 
and state-of-the-art performance~\cite{somnath_learning_2021,chen_g2retro_2023}. %
We particularly focus on their synthon completion step, because reaction centers can be  
predicted very accurately~\cite{chen_g2retro_2023}, 
but synthon completion is often more complicated~\cite{michael2017}.

Specifically, \method assigns one agent to each synthon, all of which complete the synthons by conducting 
actions step by step in a synchronized fashion. All the agents share the same action selection policy and 
select the optimal actions with full observation of other agents' states. 
\method learns the policy from offline training episodes, and %
augmented training data generated through online interactions.
The augmented data introduce new reaction patterns not included in training data, and thus allow 
\method to explore new reaction spaces. 
\method uses a reward function to evaluate the likelihood of the predicted reactants in synthesizing a product, 
and thus guides the action search. 
We compare \method with state-of-the-art synthon completion methods.
Our results demonstrate that \method can outperform these methods
with improvements as high as 14.9\%, highlighting its potential in synthesis planning.
To the best of our knowledge, \method is the first reinforcement learning method for 
synthon completion.
Our code is available at \mbox{\url{https://github.com/ninglab/RLSynC}}.
\section{Related Work}
\label{sec:related_work}

\subsection{Retrosynthesis}
\label{sec:related_work:retro}

Deep-learning-based retrosynthesis methods can be categorized into three groups: template-based, template-free and semi-template-based.
Template-based methods~\cite{dai_retrosynthesis_2019,seidl_improving_2022,sacha_molecule-edit_2023} 
use reaction templates extracted from known reactions to transform a product directly into reactants, and thus are 
 limited to reactions covered by the templates.
Template-free methods~\cite{seo_gta_2021,tu_permutation_2022,zhong_root-aligned_2022,wan_retroformer_2022} 
typically utilize the sequence representation of molecules (SMILES) and employ Transformer models to translate product SMILES strings into reactant SMILES strings, without using reaction templates. 
For example, \rsmiles~\cite{zhong_root-aligned_2022} 
uses a Transformer
to decode the reactant SMILES strings from the product SMILES strings.
\retroformer~\cite{wan_retroformer_2022} is a template-free retrosynthesis method.
\retroformer~\cite{wan_retroformer_2022} embeds both the SMILES strings and molecular graphs of products, and uses
the embeddings to predict reaction center regions and generate reactant SMILES strings.
However, these methods may generate  SMILES strings that violate SMILES grammars or chemical rules.

Semi-template-based methods~\cite{yan2020retroxpert,shi_graph_2020,wang_retroprime_2021,somnath_learning_2021,chen_g2retro_2023} 
have two steps: (1) they first identify the reaction centers and
break the product into synthons using reaction centers; and then (2) they complete synthons into
reactants.
\graphretro~\cite{somnath_learning_2021} predicts reaction centers by learning from the molecular graphs of products, and then completes synthons by classifying the subgraphs
based on whether they can realize the difference between synthons and reactants.
\gtworetro~\cite{chen_g2retro_2023} also predicts reaction centers from molecular graphs, and then completes synthons by sequentially adding rings or bonds.
\retroprime~\cite{wang_retroprime_2021} is a semi-template based retrosynthesis method.
\retroprime~\cite{wang_retroprime_2021} employs two Transformers to first translate the SMILES strings of products to synthons, and then synthons to reactants.

Recent works have leveraged structural information and attention mechanisms to perform semi-template-based and template-free retrosynthesis.
{\gtwogt}~\cite{lin_g2gt_2023} leverages a graph-to-graph transformer to transform the product into predicted reactants directly,
combining the richness of graph-based representation with the versatility of template-free methods.
{\retroexplainer}~\cite{wang_retrosynthesis_2023} is a semi-template-based method, 
which leverages graph transformers,
contrastive learning, and multi-task learning to enhance its predictive capability.
{\mars}~\cite{liu_mars_2022} uses graph attention to predict the sequence of edits and motifs required to transform the product into reactants.
They differentiate their approach by the use of chemical motifs, which are larger than atoms but smaller than leaving groups.

Existing work primarily uses exact match accuracy@$N$ to evaluate single-step retrosynthesis.
However, exact match accuracy does not account for synthetically viable predictions not matching the ground truth.
Schwaller et al. \cite{schwaller_predicting_2020} propose using forward synthesis models to evaluate single-step retrosynthesis techniques.
Similarly, we use forward synthesis for evaluation with \map, \ndcg, and \diversity.
However, \method also uses forward synthesis in training, allowing \method to explore new reaction patterns.

\subsection{Reinforcement Learning}
\label{sec:related_work:rl}

Deep reinforcement learning methods have been developed to design new small molecules. 
For example, 
{\gcpn}~\cite{you_graph_2019} uses a graph convolutional policy network to sequentially add new atoms and 
bonds to construct new molecules.
{\moldqn}~{\cite{zhou_optimization_2019}} uses
Morgan fingerprints~\cite{morgan_generation_1965} to represent molecules, and learns a deep Q-network 
to guide the addition or change of atoms and bonds,
modifying molecules to have desired properties.  
Reinforcement learning has also been applied for biological sequence design.
For example,
{\dynappo}~\cite{angermueller_dynappo_2020} uses a model-based variant of proximal-policy optimization  to generate 
DNA and peptide sequences with desired properties.
{\tcrppo}~\cite{chen_tcrppo_2023} learns a mutation policy to mutate sequences of T-cell receptors to recognize specific peptides.

Reinforcement learning has been used for multi-step retrosynthetic planning,
which seeks to find an optimal sequence of multiple reactions to synthesize a product. 
For example, Schreck \emph{et al.}~\cite{schreck_learning_2019} trains an agent to select reactions from a list  
to construct the sequence backward starting from the product, 
until all the reactants of the first reaction in the sequence
are purchasable.
{\grasp}~\cite{yu_grasp_2022} trains a goal-driven actor-critic policy to select reactions
from a tree of possible reactions until it reaches purchasable reactants in the leaf nodes of the tree.
In these methods,
reinforcement learning is used to select reactions rather than predicting reactions. 
However, there is very limited work applying reinforcement learning to retrosynthesis.
{\rcsearcher}~\cite{lan_rcsearcher_2023} applies a deep $Q$ network
to search a molecular graph for reaction centers. 
In contrast, \method uses reinforcement learning for synthon completion.

Recent efforts~\cite{nair_awac_2021,zhang_policy_2023} in reinforcement learning 
leverage offline-online learning to accelerate or improve learning.
{\awac}~\cite{nair_awac_2021} is an extension of actor-critic learning that combines prior offline
demonstration with online experience to accelerate the learning of robotics tasks.
{\pex}~\cite{zhang_policy_2023} constructs a policy set to combine offline and online policies.
{\pex} freezes and retains the offline policy during online training and
samples actions from the offline and online policies, allowing it to benefit from the offline experience without overwriting it.
These works emphasize the usefulness of combining
offline data with online experience.
Inspired by this, \method also leverages offline data to accelerate learning and
online interactions to explore the retrosynthesis prediction space.

\section{Definitions and Notations}
\label{sec:def}

A synthesis reaction involves a set of reactants $\{\reactant_i\}$ and a product molecule \product that is synthesized from these reactants.
Each reactant $\reactant_i$ has a corresponding synthon $\synthon_i$, which represents the
substructures of $\reactant_i$ that appear in \product. 
The connection point of these synthons to form the product, typically a bond, is referred to 
as the reaction center. 
Fig.~\ref{fig:retro} presents the retrosynthesis process. 
In retrosynthesis, 
a typical semi-template-based method 
first identifies the reaction center and thus the 
corresponding synthons $\{\synthon_i\}$, and then completes the synthons back to reactants $\{\reactant_i\}$. 

To complete $\synthon_i$ to $\reactant_i$, atoms may be added to $\synthon_i$  one at a time, through establishing new bonds. 
In each step $t$, the intermediate molecular structure generated from $\synthon_i$ is denoted as $\synthon_i^t$. 
With abuse of terms, such intermediate molecular structures are referred to as \emph{current} synthons.
In this manuscript, we focus on the reactions with only two reactants, because this is the most common case in synthesis reactions~{\cite{michael2017}. 
In this manuscript, the two terms ``pair of reactants'' and ``reaction''
for a product are used interchangeably, 
when no ambiguity is raised; 
the term ``prediction'' refers to the prediction of the two reactants of a product.  
Table~\ref{table:definitions} presents the key definitions and notations. 

\method is employed under the assumption that reaction centers are pre-determined or can be accurately predicted. 
This is because the potential reaction centers are typically limited, especially in the case of small molecules.
According to Chen \emph{et al.}~\cite{chen_g2retro_2023}, 
reaction center prediction can achieve as high as 97.2\% accuracy. 
However, synthon completion is often more complicated and can be realized in a variety of ways~\cite{michael2017}.
However, {\method} can be generalized to one-reactant 
or multi-reactant %
cases by having one or multiple agents, given how {\method} completes synthons (Section~\ref{sec:learning}).

\begin{table}[t]
	\centering
	\caption{Key Notations}
	\label{table:definitions}
	\resizebox{\columnwidth}{!}{
	\begin{tabular}{lp{0.87\columnwidth}} 
		\toprule
		Notation & Meaning                \\              
		\midrule                                                      
		\product                      & product molecule       \\
		$(\reactant_1,\reactant_2)$   & a pair of reactants    \\

		$(\synthon_1,\synthon_2)$                 & a pair of synthons \\
		$(\synthon^t_1,\synthon^t_2)$ & a pair of \textit{current} synthons at step $t$ \\
		\agent & an agent \\
		$t$/$T$                           & time step/time step limit \\
		\MDP & Markov decision process: $\MDP = \{\StateSpace, \ActionSpace, \Transition, \Reward\}$\\
		\StateSpace/$\state_t$ & State space/a state at step $t$ \\
		\ActionSpace/$\action_1^t$/$\action_2^t$   & Action space/an action used to update $\synthon^{t}_1/\synthon^{t}_2$ at step $t$ \\
		$A_i^t$ & the set of feasible actions for $\agent_i$ at time step $t$\\
		$\Transition$ & State transition function\\
		$\Reward(s_T)$                & reward function for terminal states\\
		\bottomrule
	\end{tabular}
S	}
\end{table}

\section{Methods}
\label{sec:methods}

\subsection{\method Model}
\label{sec:method:mode}

\method assigns one agent, denoted as \agent, 
to each synthon and uses the agent to transform its synthon into 
a reactant through a sequence of actions, assuming the reaction center is known. 
This transformation is achieved through a Markov Decision Process (MDP), denoted as 
\mbox{$\MDP = \{\StateSpace, \ActionSpace, \Transition, \Reward\}$}, including  a state space \StateSpace, 
an action space \ActionSpace, a transition function \Transition and a reward function \Reward. 

\subsubsection{State Space (\StateSpace)}
\label{sec:method:state}

\method has a discrete state space \StateSpace describing the \MDP status. 
Each state $\state_{t} \in \mathcal{S}$ is represented as:
\begin{equation}
s_{t, {\scriptsize{\product}}} = \{\synthon_{1}, \synthon_{2}, \synthon_1^t, \synthon_2^t, \product, T-t\}, 
\end{equation}
where $t$ denotes the steps of actions (Section~\ref{sec:method:action}); 
$\synthon_1$ and $\synthon_2$ are the synthons from \product that are assigned to agent \agentOne and \agentTwo, respectively;
$\synthon_1^t$ and $\synthon_2^t$ are the \emph{current} synthons generated from 
$\synthon_1$ and $\synthon_2$ after $t$ ($t=0,..., T$) steps of actions by {\agentOne} and {\agentTwo}, respectively
($\synthon_1^0 = \synthon_1$, $\synthon_2^0 = \synthon_2$); 
\product is the product molecule; and $T$ is the step limit and thus  $\state_T$ is a terminal state.
The current synthons $(\synthon_1^T, \synthon_2^T)$ in $\state_T$ are the predicted reactants. 
In \method, $T$ is set to 3 because 89.10\%
of the synthons in the benchmark USPTO-50K dataset can be completed with the addition of up to 3 atoms. 
Increasing $T$ would not significantly increase coverage of the ground truth reactants,
(e.g., $T=6$ will cover only an additional 2.96\% of the synthons in the 
benchmark dataset).
When no ambiguity is raised, $\state_{t, {\scriptsize{\product}}}$ is represented as $\state_t$ with \product dropped.

\subsubsection{Action Space (\ActionSpace)}
\label{sec:method:action}

\method has two types of actions in its action space \ActionSpace: 
(1) adding atoms via bonds, denoted as \ADD, and (2) no operation (i.e., doing nothing), denoted as \NOOP. 
For \ADD, \method allows 12 types of atoms (B, C, N, O, F, Si, P, S, Cl, Se, Br, and I) 
via single, double or triple bonds, and thus 36 types of additions. 
These additions are sufficient to complete 98.42\% of the synthons in those two-synthon cases in the benchmark
data within 3 steps. 
Adding more atom and bond types offers very little additional coverage,
but expands the action space. %
Thus, the action space is denoted as follows:
\begin{equation}
\ActionSpace=\{\ADD_1, \ADD_2, \cdots, \ADD_{36}, \NOOP\},  
\end{equation}
where each $\ADD_i$ corresponds to a specific atom type and bond type combination. 
The atom additions have to satisfy the following constraints: 
\begin{enumerate}[label={\alph*)}]
	\item The new atoms are only added to the reaction centers or atoms that are added through the previous actions; 
	\item The bonds connecting the new added atoms and the current synthons obey %
	structural or valency rules;
	\item The types of these new bonds exist in the training data.
\end{enumerate}
At each step $t$ ($t=0, ..., T$), each agent $\agent_i$ selects an action $\action_i^t$ from \ActionSpace, and applies the action to its current synthon $\synthon_i^t$
($i\!\!=\!\!1, 2$). The two agents act in a synchronized fashion and start the next step $t+1$ only when both finish step $t$. 
Note that each agent at step $t$ has perfect observations of its own action and current synthon, 
and also the other agent's current synthon. 
This full observation allows the agents to share the same policy without exchanging information, 
and thus simplifies the policy learning (Section~\ref{sec:method:policy}). 

\subsubsection{Transition Function (\Transition)}
\label{sec:method:transition}

The transition function $\Transition(\state_{t+1}|\state_{t}, \{\action_1^t, \action_2^t\})$ in \method
calculates the probability of \MDP transitioning to state $\state_{t+1}$,
given the state $\state_{t}$
and actions $\{\action_1^t, \action_2^t\}$ at step $t$. 
In \method, \Transition is deterministic, that is, \mbox{$\Transition(\state_{t+1}|\state_{t}, \{\action_1^t, \action_2^t\})=1$}. 

\subsubsection{Reward Function (\Reward)}
\label{sec:method:reward}

\method uses a final binary reward to guide its agents. 
At the terminal step $T$,
if the predicted reactants $\synthon_1^T$ and $\synthon_2^T$ exactly match the reactants given for the product \product in the training data, 
\Reward gives $\state_T$ a reward 1. 
Otherwise, a standalone forward synthesis prediction model  
is applied to predict the products that can be synthesized from $\synthon_1^T$ and $\synthon_2^T$. 
If \product is among the top-5 predictions by this model, \Reward gives $\state_T$ a reward 1; otherwise, reward 0. 
\method uses Molecular Transformer~\cite{schwaller_molecular_2019} as the forward synthesis prediction model, because it is the state of the art and achieves very high accuracy for forward synthesis prediction~\cite{JaumeSantero2023}. 
Please note that Molecular Transformer is trained from a separate dataset with no overlap with 
the training data for \method, eliminating the possibility of data leakage and bias in calculating 
rewards. 
However, \method is not bound to Molecular Transformer and can be easily adapted to any other forward synthesis prediction models for \Reward. 
Predicted reactants that receive positive rewards are referred to as \emph{correct} predictions. 

\subsubsection{State-Action Representation}
\label{sec:method:embedding}

The state-action pairs will be used to learn a state-action $Q$-value function (discussed later in Section~\ref{sec:learn:offline}). 
\method represents a state-action pair $(\state_t, \{\action_1^t, \action_2^t\})$ as follows:
\begin{equation}
\label{eqn:pairs}
\statemb_{i, t} =  \mathbf{m}_i \oplus \mathbf{m}_{j} \oplus \mathbf{m}_i^{t+1} \oplus \mathbf{m}^{t+1}_j \oplus \mathbf{p} \oplus [T-(t+1)],
\end{equation}
where $i=1,2$ indexing the agent of interest, $j=3-i$ indexing the other agent; $\mathbf{m}$'s are the Morgan fingerprint vectors 
for the corresponding 
synthons \synthon's ($\synthon_i^t$ will be transformed to $\synthon_i^{t+1}$ by $\action_i^t$, 
and $\synthon_i^{t+1}$ is represented by $\mathbf{m}_i^{t+1}$);
$\mathbf{p}$ is the Morgan fingerprint vector for the product \product; and $\oplus$ is the concatenation operation. 
The use of Morgan fingerprints is inspired by Zhou \emph{et al.}~\cite{zhou_optimization_2019}.
Morgan fingerprints~\cite{morgan_generation_1965} 
capture molecular substructure information, 
are easy to construct and do not require representation learning.

\subsection{\method Training and Prediction}
\label{sec:learning}

\subsubsection{Offline Training}
\label{sec:learn:offline}

At a state $\state_t=\{\synthon_1, \synthon_2, \synthon^t_1, \synthon^t_2, \product, T-t\}$, 
\method uses a state-action value function $Q_{\Theta}{(\state_t, \{\action^t_1, \action_2^t\})}$, parameterized by $\Theta$, to 
estimate the future rewards of $\state_t$ if the actions $\action^t_1$ and $\action^t_2$ are applied by $\agent_1$ and $\agent_2$
on $\synthon^t_1$ and $\synthon^2_2$, respectively.
$Q_{\Theta}(\state_t, \{\action^t_1, \action_2^t\})$ is modeled as a multi-layer fully-connected neural network, 
with $(\state_t, \{\action^t_1, \action_2^t\})$ represented as $\statemb_{i,t}$ (Equation~\ref{eqn:pairs}) as input to the neural network.  

\paragraph{\mbox{Offline Training Episode Generation}}

\method uses offline data of pre-computed episodes to learn $Q_{\Theta}$. 
An episode refers to a trajectory from $\state_0$ to $\state_T$ for a product \product, that is, 
\[
(\state_0, \{\action^0_1, \action^0_2\}, \state_1, \{\action^1_1, \action^1_2\}, \state_2, ..., \{\action^{T-1}_1, \action^{T-1}_2\}, \state_T, \Reward(\state_T))_{\scriptsize{\product}}.
\]
\method computes all \emph{true} episodes from training data that include known reactions to synthesize given products. 
For each product in the training set, \method also computes 4
\emph{random} episodes from a set of 
random reactions that are not included in the training data. 
These random reactions are generated by 
taking random actions on the synthons of the products in the training data, and their rewards are calculated using \Reward.  
While all the \emph{true} episodes for known reactions have positive rewards, most episodes for random reactions have zero rewards.
By training on both positive- and zero-reward episodes, the agents can learn actions to take as well as actions to avoid, thereby improving their overall performance.

\paragraph{$Q$-Value Function Learning}

From the offline training episodes, 
\method uses a SARSA~\cite{sutton2018}-like approach to approximate the Q-value $\tilde{Q}(\state_t, \{\action^{t}_1, \action_2^t\})$ 
as follows:
\begin{equation}
\label{eqn:q}
\!\!\!\!\tilde{Q}(\state_t,\{\action_1^{t}, \action_2^t\})\!=\!\ \!\!
    \begin{cases}
    	\gamma \tilde{Q}(\state_{t+1}, \{\action_1^{t+1}, \action_2^{t+1}\})\!&\!\!\!\text{if $t\!<\!T\!-\!1$, }\\
    	\Reward(\state_{t+1})\!&\!\!\!\text{if $t\!=\!T\!-\!1$, }\\
    \end{cases}
\end{equation}
where $0<\gamma<1$ is the discount factor and \Reward is the reward function.
With the approximate Q-values, \method learns $Q_{\Theta}(\state_t, \{\action^t_1, \action^t_2\})$ by minimizing the following loss function: 
\begin{equation}
\label{eqn:loss}
\mathcal{L}({\Theta}|\data)\!=\!\frac{1}{|\data|}
\sum_{\mathclap{\scriptsize{\quad\quad\quad\quad\quad{\{\state_t, \{\action^t_1, \action^t_2\}\}\in \data}}}}
(Q_\Theta(\state_t, \{\action^t_1, \action^t_2\}) - q_t)^2 + \alpha \mathcal{R}_{\ell_2}({\Theta}), 
\end{equation}
where \data is the set of all $\{\state_t, \{\action^t_1, \action^t_2\}\}$ pairs 
from the offline training episodes and $|\data|$ is its size; $q_t = \tilde{Q}(\state_t, \{\action^t_1, \action^t_2\})$ is calculated from Equation~\ref{eqn:q}; 
 $\mathcal{R}_{\ell_2}$ is the $\ell_2$-norm regularizer over $\Theta$; and $\alpha > 0$ is a trade-off parameter.

\subsubsection{Training Data Augmentation via Online Interactions}
\label{sec:learn:data}

The offline training has the advantage of sourcing experience from known reactions, and thus exposing \method to successful
synthon completion patterns (i.e., the \emph{true} episodes). 
However, it also eliminates the opportunity for \method to explore and discover new synthon completion patterns 
that are not included in the training data.
To mitigate this issue, inspired by Nair \emph{et al.}~\cite{nair_awac_2021}, \method deploys an innovative strategy to augment the offline training data
via online interactions, denoted as \augalg. 

\paragraph{\mbox{Augmentation through Greedy Action Selection}}
\label{sec:learn:data:top1}

\method first learns $Q_{\Theta}$ using the known and 
random reactions as discussed above (Section~\ref{sec:learn:offline}). 
The optimized agents are then applied to the original unaugmented training data (only 
known reactions, excluding random reactions), 
and generate one new episode for each unique product in the training data
 from the actions they take and the rewards the actions receive (action selection discussed later in Section~\ref{sec:method:policy}).
These new episodes are added to and augment the training data. 
$Q_{\Theta}$ is then re-learned from this new augmented training data and 
the new agents generate another set of new episodes for further training data augmentation. 
The above process is iterated until the performance of the agents on the separate validation set is not improved. 

\paragraph{Augmentation through Top-$N$ Prediction Search}
\label{sec:learn:data:topN}

Once the agents' performance is stabilized,
a more aggressive strategy is deployed to generate additional new episodes through 
top-$N$ prediction search (discussed
in Section~{\ref{sec:learn:topn}}) and 
tracking back the episodes leading to the top-$N$ predictions, resulting in $N$ episodes for each product. 
Multiple iterations of such top-$N$-prediction-based new episode generation are 
conducted and 
training data is augmented until validation set performance ceases to improve.  
A final $Q_{\Theta}$ will be learned over the final augmented training data. 

Augmenting the training data with additional positive-reward episodes exposes \method to new reactions that  do not exist in the training data, allowing \method to explore beyond the limits of the data.
Augmenting the training data with additional zero-reward episodes helps \method correct inaccurate $Q$-value estimations, enhancing its robustness and effectiveness.
Algorithm~\ref{alg:augment} presents \augalg.

\begin{algorithm}[h]
	\caption{\augalg}
	\floatname{algorithm}{Procedure}
	\renewcommand{\algorithmicrequire}{\textbf{Input:}}
	\renewcommand{\algorithmicensure}{\textbf{Output:}}
	\newcommand{\dataset}[1]{$\mathcal{E}_\text{#1}$} %
    	\label{alg:augment}
	\begin{algorithmic}[1]
	\REQUIRE \trainset, \randomset, \validationset, $k$, $N$
	\ENSURE $Q_{\Theta^*}$
	\STATE \trainaug = $\trainset \cup \randomset$	\COMMENT{$\mathcal{E}$ refers to the set of episodes}
	\STATE $\mathcal{E}^*_{\text{aug}} = \trainaug$
	\STATE $\Theta = \text{arg}\min_{\Theta} \mathcal{L}({\Theta}|\mathcal{E}^*_{\text{aug}})$ \COMMENT{Equation~\ref{eqn:loss}}
	\STATE $\text{map}^*= -1$, flag = 0
	\STATE $\text{map} = \map(Q_{\Theta}, \validationset)$ \COMMENT{Equation~\ref{eqn:map}}
        \WHILE{$\text{map} > \text{map}^*$}
		\STATE $\text{map}^* = \text{map}$, $\Theta^* = \Theta$,  \trainaug = $\mathcal{E}^*_{\text{aug}}$

	         \FORALL{\product in \trainset}
			\IF{flag == 0}
			 	\STATE {$\mathcal{E}^*_{\text{aug}}\!\!\!=\!\! \trainaug \cup \topsearch(\synthon_1, \synthon_2, \product, T, 1, 1, Q_{\Theta})$}
			\ELSE
				\STATE \mbox{$\mathcal{E}^*_{\text{aug}}\!\!= \trainaug \cup \topsearch(\synthon_1, \synthon_2, \product, T, k, N, Q_{\Theta})$}
			\ENDIF
		\ENDFOR	

		\STATE $\Theta = \text{arg}\min_{\Theta} \mathcal{L}({\Theta}|\mathcal{E}^*_{\text{aug}})$ \COMMENT{Equation~\ref{eqn:loss}}
		\STATE $\text{map} = \map(Q_{\Theta}, \validationset)$  \COMMENT{Equation~\ref{eqn:map}}
		\IF{$\text{map} \le \text{map}^*$ \AND flag == 0}
			\STATE flag = 1, $\text{map}^*$ = -1, $\Theta = \Theta^*$,  $\mathcal{E}^*_{\text{aug}}$ = \trainaug
		\ENDIF
	\ENDWHILE
	\RETURN $Q_{\Theta^*}$
    \end{algorithmic}
\end{algorithm}

\subsubsection{{Action Selection Policy}}
\label{sec:method:policy}

In predicting the reactants of a new product \product (i.e., in validation set or test set), 
the agents of \method predict and select actions
to complete synthons via up to $T$ steps. 
To predict the action $\action_i^t$ for agent $\agent_i$ at time $t$, \method first identifies 
the set of all chemically feasible atom 
additions, denoted as $A_i^t$ ($A_i^t  \subseteq \ActionSpace$), for $\synthon^t_{i}$ satisfying the constraints as 
in Section~{\ref{sec:method:action}}.
$\agent_i$ samples the best action $\action_i^t \in A_i^t$ ($i=1,2$) if the action has the maximum predicted $Q$-value, as follows:
\begin{equation}
\label{eqn:maxq1}
\vspace{-5pt}
\action^{t}_1  = \textstyle{\arg \max_{\scriptsize{\action\in A_1^t}} Q_{\Theta}(\state_t, \{\action, \NOOP\})},
\end{equation}
\begin{equation}
\label{eqn:maxq2}
\action^{t}_2  = \textstyle{\arg \max_{\scriptsize{\action\in A_2^t}} Q_{\Theta}(\state_t, \{\NOOP, \action\})}.
\end{equation}
That is, each agent assumes \NOOP from the other agent, but observes the other agent's 
\emph{current} synthon in $\state_t$ in order to select its next optimal action. 
Please note that the two agents share the same $Q_{\Theta}$ function
and follow the same policy. %

\subsubsection{Top-$N$ Prediction Search}
\label{sec:learn:topn}

\method uses a novel greedy search algorithm, denoted as \topsearch, 
to identify the top-$N$ predicted reactions with the highest $Q$-values. 
In training data augmentation, such top-$N$ pairs will be used to augment the training data (Section~\ref{sec:learn:data:topN}). 
In predicting reactants for new products (e.g., in test set), 
the top-$N$ predictions provide more options for synthesis planning. 
Specially, at step $t=0$, instead of selecting only one action, each agent selects $k$ actions with the top-$k$ highest $Q$-values calculated based on Equation~\ref{eqn:maxq1} and~\ref{eqn:maxq2}
(i.e., instead of ``$\max$'' in Equation~\ref{eqn:maxq1} and~\ref{eqn:maxq2}, use ``top-$k$''). 
Thus, there will be $k^2$ possible next states,
resulting from all possible action combinations from the two agents. 
In each of the possible next states, each agent selects again $k$ actions for its current synthon. 
Through $T$ steps, this process will result in $k^{2T}$ predicted reactions at the terminal state. 
\method sorts all these predicted reactions using their $Q$-values and selects the top-$N$ predictions. 
While this process can be expensive, the actions on each of the possible next states can be done independently, and 
this process can be implemented in parallel. 
In our experiments, we used $k=3$ and $T=3$, and therefore the computational 
cost is low.
The algorithm is presented in Algorithm~\ref{alg:search}.

\begin{algorithm}[h]
	\caption{\topsearch}
	\floatname{algorithm}{Procedure}
	\renewcommand{\algorithmicrequire}{\textbf{Input:}}
	\renewcommand{\algorithmicensure}{\textbf{Output:}}
	\begin{algorithmic}[1]
		\label{alg:search}
		\REQUIRE 
		$\synthon_1$, $\synthon_2$, \product, $T$, $k$, $N$, $Q_\Theta$
		\ENSURE Top-$N$ predictions $\data_{\text{top-}N}$\\
		\STATE $\synthon_1^0 = \synthon_1$, $\synthon_2^0 = \synthon_2$\\
		\STATE $s_0 = \{\synthon_1, \synthon_2, \synthon_1^0, \synthon_2^0, \product, T\}$, ~$\mathbb{S}_0 = \{s_0\}$ \\
		\FOR{t = 1 \TO T}
			\STATE $\mathbb{S}_t = \emptyset$  \\
			\FORALL{$s_{t-1} \in \mathbb{S}_{t-1}$}
				\FORALL{agent $\agent_i$ ($i=1,2$)}
					\STATE $V_i = \emptyset$ \\
					\FORALL{$\action \in A_i^{t-1}$}
						\STATE $V_i = V_i \cup \{(\action: Q_\Theta(s_{t-1}, \{\action, \NOOP\}))\}$
						\COMMENT{or $\{\NOOP, \action\}$, Equation~\ref{eqn:q}}
					\ENDFOR
					\STATE $V_i = \text{sorted}(V_i, \text{`decreasing'})$[1:$k$]
				\ENDFOR
				\FORALL{$\{\action_1^{t-1}, \action_2^{t-1}\} \in V_1 \times V_2$}
					\STATE identify $s_t$ s.t. $\Transition(s_t|s_{t-1}, \{\action_1^{t-1}, \action_2^{t-1}\}) = 1$
					\STATE $q_t = Q_\Theta(s_{t-1}, \{\action_1^{t-1}, \action_2^{t-1}\})$
					\STATE $\mathbb{S}_t = \mathbb{S}_t \cup \{\{(\action_1^{t-1}, \action_2^{t-1}), \state_t\}:q_t$\}
				\ENDFOR
			\ENDFOR
		\ENDFOR
		\RETURN $\data_{\text{top-}N} = \text{sorted}(\mathbb{S}_T, \text{`decreasing'})$[1:$N$]
	\end{algorithmic}
\end{algorithm}

\section{Experimental Settings}
\label{sec:settings}

\subsection{Data}
\label{sec:settings:data}

We use the benchmark USPTO-50K dataset~\cite{schneider_whats_2016} in our experiments, which contains 50,016 chemical reactions.
We use the same training, validation and testing division as in the literature~\cite{yan2020retroxpert},
resulting in 40,008 reactions for training, 5,001 for validation, and 5,007 for testing.
From each set, we use only the reactions that satisfy the following constraints to train/evaluate \method:
\begin{enumerate}[label={\alph*)}]
	\item The reaction has exactly two reactants;
	\item The synthons can be completed to the ground-truth reactants by adding no more than three atoms.
\end{enumerate}
After applying the above filter, our training, validation, and test sets contain
25,225, 3,172 and 3,167 reactions, respectively.
The baseline models are trained using all 40,008 training reactions and tested on the same test set as \method.

\subsection{Baselines}

We compare \method against three state-of-the-art retrosynthesis methods,
including 
\graphretro~\cite{somnath_learning_2021},
\gtworetro~\cite{chen_g2retro_2023}, and
\rsmiles~\cite{zhong_root-aligned_2022}.
In the experiments of synthon completion (Section~\ref{sec:results:synthon}), we compare \method 
against the synthon completion components of \gtworetro, \graphretro, and the semi-template-based version of \rsmiles, referred to as \RPtoStoR,
because, as semi-template-based methods, they have a synthon completion component that can be used separately.
Each of these methods takes a different approach to retrosynthesis prediction:

\begin{itemize}
	\item{ \graphretro \cite{somnath_learning_2021} uses graph message passing networks to perform semi-template-based retrosynthesis.
		   \graphretro treats reaction center prediction as a set of bond classification problems.
		   For synthon completion, \graphretro selects leaving groups to append from a limited set of options.  }
	\item{ 
		\gtworetro \cite{chen_g2retro_2023} also uses graph message passing networks to perform semi-template-based retrosynthesis.
		\gtworetro learns a set of atom and bond regression problems to score reaction centers.
		\gtworetro completes synthons by performing a sequence of additions to the molecule.
	}
	\item{
		\rsmiles \cite{zhong_root-aligned_2022} treats retrosynthesis as a sequence-to-sequence prediction task and
		uses a Transformer~\cite{vaswani_attention_2017} model trained
		 on root-aligned SMILES.
		By aligning the training set SMILES, \rsmiles minimizes the edit distance
		between the product and reactant strings,
		making it easier for the model to learn accurate predictions.
		\rsmiles is also able to augment its training data with multiple
		SMILES aligned to different root atoms.
		\rsmiles provides both template-free and semi-template-based versions.
		
		\begin{itemize}
		\item \RPtoR: The template-free version of \rsmiles
		directly outputs the reactant SMILES based on the product SMILES input.
		\item \RPtoStoR: The semi-template-based version of \rsmiles uses two sequence-to-sequence models:
		the first produces synthons given a product, and the second produces reactants given the synthons.
		\end{itemize}
	}
\end{itemize}

\subsection{Evaluation Metrics}
\label{sec:settings:metrics}

We evaluate different methods in terms of the correctness, diversity, and validity of their predicted reactants.
Consistent with common practice in reinforcement learning, we use the reward function in these metrics.
Please note that the reward function's forward synthesis model is standalone; it remains fixed while training \method.
During testing, \method does not use the reward function to predict the reactants, but uses the learned $Q_\Theta$ to guide its prediction.

\subsubsection{Correctness Metrics}

Two predicted reactants of a product are considered \emph{correct} if they receive reward 1
(Section~\ref{sec:method:reward}). To measure the 
correctness of top-$N$ predicted reactant pairs, we use mean average precision at top-$N$ (\map).
We define \map as follows:
\begin{equation}
\label{eqn:map}
\map =  \frac{1}{|\testset|}\sum_{\scriptsize{\product \in \testset}} 
	\frac{1}{N}\sum_{k=1}^{N} \Reward(\{\synthon_1^T, \synthon_2^T\}_{\scriptsize{\product, k}}),  
\end{equation}
where \testset is the test set, \product is a product in \testset, \Reward is the reward (Section~\ref{sec:method:reward}), 
$\{\synthon_1^T, \synthon_2^T\}_{\scriptsize{\product, k}}$ is the 
$k$-th ranked predicted reactants (i.e., at the terminal step $T$, see Section~\ref{sec:method:state}) for \product. 
Higher \map indicates better correctness among top-$N$ predictions. 
Note that \map is different from accuracy@$N$ in retrosynthesis prediction~\cite{somnath_learning_2021,zhong_root-aligned_2022,chen_g2retro_2023}, which compares the predictions
only with the ground-truth reactions.

We also use normalized discounted cumulative gain at top-$N$ (\ndcg)~\cite{jarvelin_cumulated_2002}, which is a popular metric in evaluating ranking. 
In our experiments, \ndcg uses the rewards as gains and captures both the rewards and the ranking positions 
of the predictions. Higher \ndcg indicates that correct predictions tend to be ranked higher. 

\subsubsection{Diversity Metrics}

A diverse set of correct predictions enables a broad range of viable options to synthesize a product, and thus is preferred in synthesis planning.
We measure diversity using the
dissimilarity among the top-$N$ correct predictions, as follows:
\begin{equation}
	\label{eqn:diversity}
	\diversity = \frac{1}{N|\testset|}\sum_{\scriptsize{\product \in \testset}}	\sum_{i=1}^{N}
		{\Reward(\reaction_i)_{\scriptsize{\product}}}\mathsf{dsim}(\reaction_i)_{\scriptsize{\product}},
\end{equation}
where $\reaction_j$ is the $j$-the ranked predicted reactions for {\product} (i.e., $\reaction_j = \{\synthon_1^T, \synthon_2^T\}_{\scriptsize{\product}, j}$); 
$\Reward(\reaction_j)_{\scriptsize{\product}}$ is the reward for $\reaction_j$ (for correct predictions, $\Reward(\reaction_j)_{\scriptsize{\product}}=1$; otherwise 0); 
$\mathsf{dsim}(\reaction_i)_{\scriptsize{\product}}$ is the minimum dissimilarity of $\reaction_i$ to higher ranked correct predicted reactions for product $\product$ as defined below:
\begin{equation*}
	\resizebox{\columnwidth}{!}{$\mathsf{dsim}(\reaction_i)_{\scriptsize{\product}}\!=\!\begin{cases} 
		0  
		&\text{if } \sum_{j=1}^{i-1} \Reward(\reaction_j)_{\scriptsize{\product}} = 0\text{ or }$i=1$,\\
		\min\nolimits_{j\in \{1, ..., i-1\}, \scriptsize{\Reward}(\reaction_j)_{\scriptsize{\product}} = 1}
		(1-\text{sim}(\reaction_i, \reaction_j)) 
		&\text{otherwise},
	\end{cases}$}
\end{equation*}
where $\text{sim}(\reaction_i, \reaction_j)$ 
is the Tanimoto similarity between two predicted reactions.
Following the definition from Chen \emph{et al.}~\cite{chen_g2retro_2023},
we define similarity between predicted pairs of reactants as follows:

\begin{equation}
\label{eqn:sim}
\begin{aligned}
\!\!\!
\text{sim}(\reaction_i, \reaction_j) 
= \frac{1}{2}\max(&\text{sim}_{\scriptsize{\synthon}}(\synthon_{1j}^T, \synthon_{1i}^T) 
								+ \text{sim}_{\scriptsize{\synthon}}(\synthon_{2j}^T, \synthon_{2i}^T), \\
								&\text{sim}_{\scriptsize{\synthon}}(\synthon_{1j}^T, \synthon_{2i}^T) + \text{sim}_{\scriptsize{\synthon}}(\synthon_{2j}^T, \synthon_{1i}^T)), 
\end{aligned}
\end{equation}
where $\synthon_{kj}^T$ is the $k$-th predicted reactant ($k=1,2$) for the \mbox{$j$-th} ranked prediction for the product $\product$;
and $\text{sim}_{\scriptsize{\synthon}}$ is the Tanimoto similarity~\cite{Bajusz2015} between two
molecules.
Note that in Equation~\ref{eqn:sim}, we assume reactions have exactly two predicted reactants.
Baseline methods can predict reactions with any number of reactants.
Again, following Chen et al. \cite{chen_g2retro_2023},
when either predicted reaction has fewer or more than two reactants, we define similarity as follows:
\begin{equation*}
	\text{sim}(\reaction_i, \reaction_j) = \text{sim}_{\scriptsize{\synthon}}(\synthon_{j}^T, \synthon_{i}^T),
\end{equation*}
where $\synthon_{i}^T$ denotes a disconnected, composite molecule consisting of all of the predicted reactants for $\reaction_i$.

\section{Results}
\label{sec:results}

We first compare \method directly to the synthon completion components of the baselines 
by providing known reaction centers to all the methods.
Please note that all the methods are tested only on reactions that have two reactants;  
\method is trained on reactions with only two reactants
(Section~\ref{sec:settings:data}); 
all the baseline models are from their original publications and implementations,  
and were trained over USPTO-50K dataset~\cite{schneider_whats_2016}
(i.e., including reactions of various numbers of reactants).

\subsection{Evaluation on Synthon Completion}
\label{sec:results:synthon}

\subsubsection{Correctness Evaluation}

\begin{table}[t]
    \centering
    \caption{\map for Synthon Completion}
    \label{tbl:map:2sync_only2}
\begin{threeparttable}
      \begin{tabular}{%
        @{\hspace{3pt}}l@{\hspace{3pt}}
        @{\hspace{3pt}}c@{\hspace{3pt}}
        @{\hspace{3pt}}c@{\hspace{3pt}}
        @{\hspace{3pt}}c@{\hspace{3pt}}
        @{\hspace{3pt}}c@{\hspace{3pt}}
        @{\hspace{3pt}}c@{\hspace{3pt}}
        @{\hspace{3pt}}c@{\hspace{3pt}}
        @{\hspace{3pt}}c@{\hspace{3pt}}
        @{\hspace{3pt}}c@{\hspace{3pt}}
        @{\hspace{3pt}}c@{\hspace{3pt}}
        @{\hspace{3pt}}c@{\hspace{3pt}}
              }\toprule
              $N$           & 1 & 2 & 3 & 4 & 5 & 6 & 7 & 8 & 9 & 10 \\
              \midrule
              \RPtoStoR      & \ul{\bf{0.953}} & 0.863        & 0.795       & 0.734       & 0.681       & 0.636       & 0.601        & 0.571        & 0.545      & 0.523 \\
            \graphretro & 0.912       & 0.861        & 0.827       & 0.804       & 0.779       & 0.747       & 0.715        & 0.687        & 0.662      & 0.643 \\
            \gtworetro  & {0.950} &  \ul{0.893} & \ul{0.855} & \ul{0.823} & \ul{0.789} & \ul{0.753} & \ul{0.720}  & \ul{0.691} & \ul{0.665} & \ul{0.645} \\
            \method     & 0.927      &  \bf{0.898} & \bf{0.874} & \bf{0.845} & \bf{0.822} & \bf{0.803} & \bf{0.784} & \bf{0.769} & \bf{0.754} & \bf{0.741} \\
            \midrule
            imprv.(\%)         & -2.7* & 0.6 &2.2* &2.7* &4.2* &6.6* &8.9* &11.3* &13.4* &14.9* \\
              \bottomrule
        \end{tabular}
        
        \begin{tablenotes}[normal,flushleft]
            \begin{footnotesize}
            \item
            The best performance for each $N$ is in \textbf{bold}, 
            and the best performance among the baseline method is \ul{underlined}. 
            The row ``imprv.(\%)” presents the
            percentage improvement of \method over the best-performing baseline methods (\ul{underlined}).
            The * indicates that the improvement is statistically significant at 95\%
            confidence level.
        \par
        \end{footnotesize}
        \end{tablenotes}
    \end{threeparttable}
\end{table}

\begin{table}[htb]
    \centering
    \caption{\ndcg for Synthon Completion}
    \label{tbl:ndcg:2sync_only2}
\begin{threeparttable}
      \begin{tabular}{%
        @{\hspace{3pt}}l@{\hspace{3pt}}
        @{\hspace{3pt}}c@{\hspace{3pt}}
        @{\hspace{3pt}}c@{\hspace{3pt}}
        @{\hspace{3pt}}c@{\hspace{3pt}}
        @{\hspace{3pt}}c@{\hspace{3pt}}
        @{\hspace{3pt}}c@{\hspace{3pt}}
        @{\hspace{3pt}}c@{\hspace{3pt}}
        @{\hspace{3pt}}c@{\hspace{3pt}}
        @{\hspace{3pt}}c@{\hspace{3pt}}
        @{\hspace{3pt}}c@{\hspace{3pt}}
        @{\hspace{3pt}}c@{\hspace{3pt}}
              }\toprule
              $N$           & 1 & 2 & 3 & 4 & 5 & 6 & 7 & 8 & 9 & 10 \\
              \midrule
              \RPtoStoR      & \ul{\bf{0.953}}  & 0.883 			& 0.831 		& 0.784 		& 0.742 		& 0.707 		& 0.678 		& 0.652 		& 0.630 		& 0.611\\
              \graphretro & 0.912 		   & 0.872 			& 0.846 		& 0.827 		& 0.807 		& 0.784 		& 0.760  		& 0.739 		& 0.719 		& 0.703 \\
              \gtworetro  & 0.950  		   & \ul{\bf{0.905}} 	& \ul{0.876} 	& \ul{0.851} 	& \ul{0.825} 	& \ul{0.798}	& \ul{0.773} 	& \ul{0.750} 	& \ul{0.729} 	& \ul{0.712} \\
              \method     & 0.927  		   & \bf{0.905} 		& \bf{0.886} 	& \bf{0.865}	& \bf{0.847} 	& \bf{0.832}	& \bf{0.817}	& \bf{0.805}	& \bf{0.793}	& \bf{0.782} \\
              \midrule
              impv.(\%) & -2.7*	& 0.0	& 1.1*	& 1.6*	& 2.7*	& 4.3*	& 5.7*	& 7.3*	& 8.8*	& 9.8* \\
              \bottomrule
        \end{tabular}
        \begin{tablenotes}[normal,flushleft]
            \begin{footnotesize}
            \item
            The annotations in this table are the same as those in Table~\ref{tbl:map:2sync_only2}.
        \par
        \end{footnotesize}
        \end{tablenotes}
    \end{threeparttable}
\end{table}

Table~\ref{tbl:map:2sync_only2} and Table~\ref{tbl:ndcg:2sync_only2} 
presents the performance of all the methods in terms of \map
and \ndcg in completing given synthons. 
As Table~\ref{tbl:map:2sync_only2} shows, in terms of \map, \method outperforms the baseline methods consistently for $N\in[2, 10]$. 
\gtworetro is the best baseline, with the best \map among all baselines at $N\in[2,10]$, and 
very close to the best MAP@1 achieved by \RPtoStoR. 
\method significantly outperforms \gtworetro on 9 results, with 
the best improvement 14.9\% %
at $N\!\!=\!\!10$, average improvement 7.2\% %
over $N\in[2, 10]$, and 8 improvements statistically significant.
As $N$ increases, the performance improvement from \method over other methods also increases, indicating that 
\method generates more positive-reward reactions than other methods among top-$N$ predictions.
This capability of \method could facilitate the design of multiple synthesis reactions in synthesis planning.
Note that as $N$ increases, top-$N$ predictions tend to include more incorrect predictions, 
leading to a decrease in \map.

Similar trends can be observed in Table~\ref{tbl:ndcg:2sync_only2}: 
In terms of \ndcg, 
\method consistently outperforms the best baseline 
method \gtworetro at $N\in[3,10]$, all with statistically significant improvement; 
the best improvement is 9.8\%
at $N = 10$, and average improvement 5.2\% over $N\in[3,10]$. 
\method achieves the same performance as \gtworetro at $N = 2$.
\RPtoStoR achieves the best NDCG@1 performance among all the methods, but its performance dramatically decreases
for larger $N$. 
\graphretro's performance is between \gtworetro and \RPtoStoR. 
The difference between \map and \ndcg is that \ndcg discounts the impact of low-ranking correct predictions.
The fact that \method achieves both high \map and \ndcg indicates that \method predicts more correct reactions at high ranks (e.g., top 2, top 3).

\gtworetro completes synthons into reactants by sequentially attaching substructures (i.e., bonds or rings) starting from the 
reaction centers.
Unlike \gtworetro, which exclusively learns from known reactions, \method benefits from random reactions and online iterations of augmented data, 
therefore, it can overcome the potential limitations and biases in the known reactions.
In addition, \method can adapt and improve from past mistakes by 
learning from online iterations of data augmentation, 
and thus correct inaccurate $Q$-value predictions in zero-reward episodes.
These advantages enable \method to finally outperform \gtworetro.

\RPtoStoR employs a Transformer to translate root-aligned synthon SMILES strings into reactant strings.
It also augments the training and test synthon SMILES strings with varied atom orders to improve the performance further.
However, \RPtoStoR is trained to recover the unique SMILES strings of ground-truth reactants in the training data, disregarding other possible reactants to synthesize the same product.
In contrast, by augmenting data with online interactions, particularly through top-$N$ search, \method 
focuses beyond just the top-1 prediction and aims to maximize the rewards for the overall top-$N$ predictions, and achieves better \map for $N\!\in\![2,10]$ than \RPtoStoR.
\graphretro formulates synthon completion as a classification problem over subgraphs. 
However, it ignores the impact of predicted subgraphs on the overall structures of resulting reactants, which may lead to incorrect predictions. 
Unlike \graphretro, \method predicts reactants by adding feasible 
bonds and atoms to \emph{current} synthons under the guidance of rewards for resulting molecules.
As a result, \method outperforms \graphretro on \map at $N\!\in\![1,10]$.

\subsubsection{Diversity Evaluation}

Table~\ref{tbl:2sync_only2:div} presents the performance of different methods in terms of \diversity
(Equation~\ref{eqn:diversity}) among their correctly predicted reactants. 
Similar trends to those for \map
and \ndcg
can be observed for \diversity. 
\method outperforms the best baseline, {\graphretro,} at {$N\in[2, 10]$},
all with statistical significance, with the best improvement {6.8\%}
at {$N=10$}, and average improvement {6.8\%} over {$N\in[2, 10]$}.

While all the baseline methods are limited by the synthon completion patterns within known reactions from the training data, 
\method is able to discover patterns not present in the training data by learning from augmented data 
via online interactions.
These newly discovered patterns could contribute to the better \diversity at {$N\in[2,10]$} for \method. 
Higher \diversity indicates a higher variety of correctly predicted reactants.
Please note that diversity in predicted reactions is always desired, as it can enable the exploration of multiple synthetic options.
This makes \method a potentially preferable tool in synthetic design.

\begin{table}[t]
	\centering
	\caption{\diversity for Correct Synthon Completion}
	\label{tbl:2sync_only2:div}
	\begin{threeparttable}
      \begin{tabular}{%
        @{\hspace{3pt}}l@{\hspace{3pt}}
        @{\hspace{3pt}}c@{\hspace{3pt}}
        @{\hspace{3pt}}c@{\hspace{3pt}}
        @{\hspace{3pt}}c@{\hspace{3pt}}
        @{\hspace{3pt}}c@{\hspace{3pt}}
        @{\hspace{3pt}}c@{\hspace{3pt}}
        @{\hspace{3pt}}c@{\hspace{3pt}}
        @{\hspace{3pt}}c@{\hspace{3pt}}
        @{\hspace{3pt}}c@{\hspace{3pt}}
        @{\hspace{3pt}}c@{\hspace{3pt}}
      }
            \toprule
            $N$         & 2 & 3 & 4 & 5 & 6 & 7 & 8 & 9 & 10 \\
            \midrule
            \RPtoStoR   & 0.129   & 0.114   & 0.104   & 0.096   & 0.089   & 0.084   & 0.080   & 0.077   & 0.072   \\
            \graphretro  & \underline{0.156}   & \underline{0.150}   & \underline{0.146}   & \underline{0.142}   & \underline{0.137}   & \underline{0.131}   & \underline{0.126}   & \underline{0.121}   & \underline{0.117}   \\
            \gtworetro   & 0.154   & 0.148   & 0.143   & 0.137   & 0.131   & 0.125   & 0.120   & 0.116   & 0.113   \\
            \method & \bf{0.164}   & \bf{0.156}   & \bf{0.154}   & \bf{0.147}   & \bf{0.142}   & \bf{0.136}   & \bf{0.131}   & \bf{0.128}   & \bf{0.125}   \\
            \midrule
            impv. & 5.1*   & 4.0*   & 5.5*   & 3.5*   & 3.6*   & 3.8*   & 4.0*   & 5.8*   & 6.8*   \\
            \bottomrule
    \end{tabular}
    \begin{tablenotes}[normal,flushleft]
        \begin{footnotesize}
        \item
        The annotations in this table are the same as those in Table~\ref{tbl:map:2sync_only2}.
        Diversity@1 is always 0 based on the definition (Equation~\ref{eqn:diversity}) so it is not presented in 
        this table. 
      \par
      \end{footnotesize}
    \end{tablenotes}
  \end{threeparttable}
\end{table}

\paragraph{Diversity in Leaving Groups}

We evaluate the ability of \method to discover leaving groups -- the difference between 
synthons and their completed reactants -- that are not present in the training data.
The ground-truth reactants in our training data set contain
35 distinct leaving groups.
\method uses 229 leaving groups in its correct
top-10 predictions on the test set. 
Among the correct top-10 predictions on the test set,
46.8\% use novel leaving groups.
This highlights the ability of \method to explore novel reaction patterns,
which can facilitate new synthesis routes in synthetic planning.

\subsection{Validity Evaluation}
\label{sec:validity}

We also evaluate the validity of the predicted reactions returned by retrosynthesis methods.
We consider a predicted reaction to be valid if all the SMILES strings for the predicted reactants are valid. 
Valid SMILES strings should obey standard valency rules.
Validity at top-$N$ %
is calculated as the percentage of valid predictions among all the top-$N$ predictions. 

We also evaluate the validity of the completed synthons by the different methods. 
\method and \gtworetro always achieve 100\% validity among its top-$N$ predictions ($N\in[1, 10]$). 
This is because they %
enforce validity (Section \ref{sec:method:action}), and
complete synthons by adding atoms and bonds obeying valency rules. 
{\RPtoStoR} can achieve on average 99.3\% validity
among top-$N$ ($N\in[1, 10]$)
predictions, and \graphretro can achieve 98.4\% validity.
{\RPtoStoR} formulates synthon completion as a sequence-to-sequence translation problem,
so it cannot guarantee valency or the validity of the output strings.
\graphretro leverages a graph-based model of molecules,
but uses a more relaxed set of valency rules when editing the molecular graphs.

\subsection{Evaluation on Data Augmentation}

\subsubsection{Correctness Evaluation}

\begin{figure}[h]
    \centering
    \begin{subfigure}{0.32\linewidth}
      \centering
      \includegraphics[width=\linewidth]{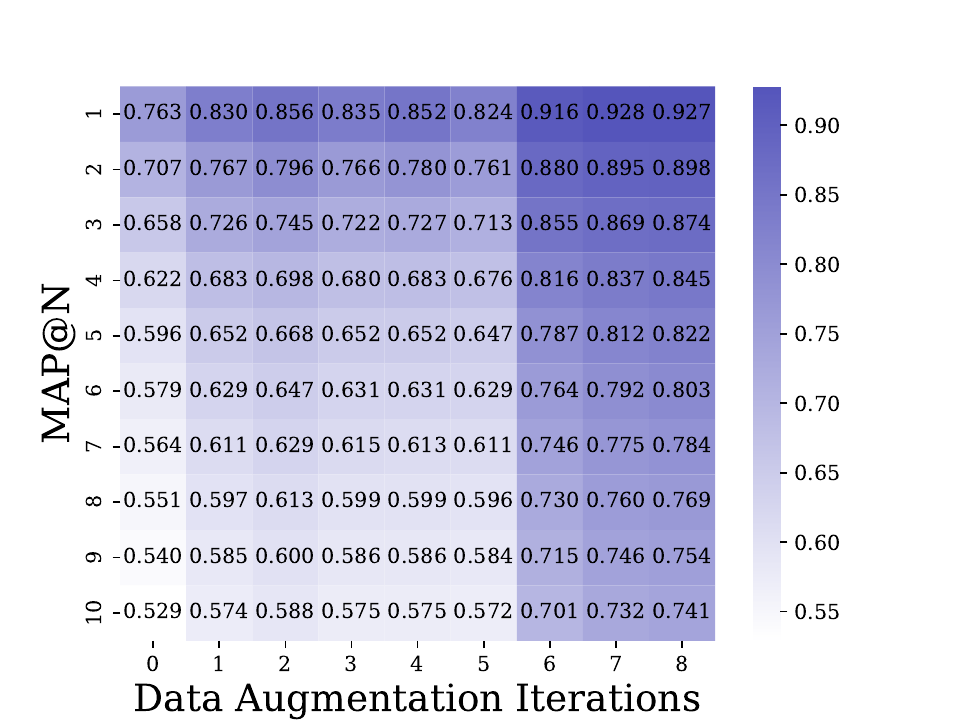}
        \caption{\map}
        \label{fig:iter_map}
    \end{subfigure}
    ~
    \begin{subfigure}{.32\linewidth}
      \centering
      \includegraphics[width=\linewidth]{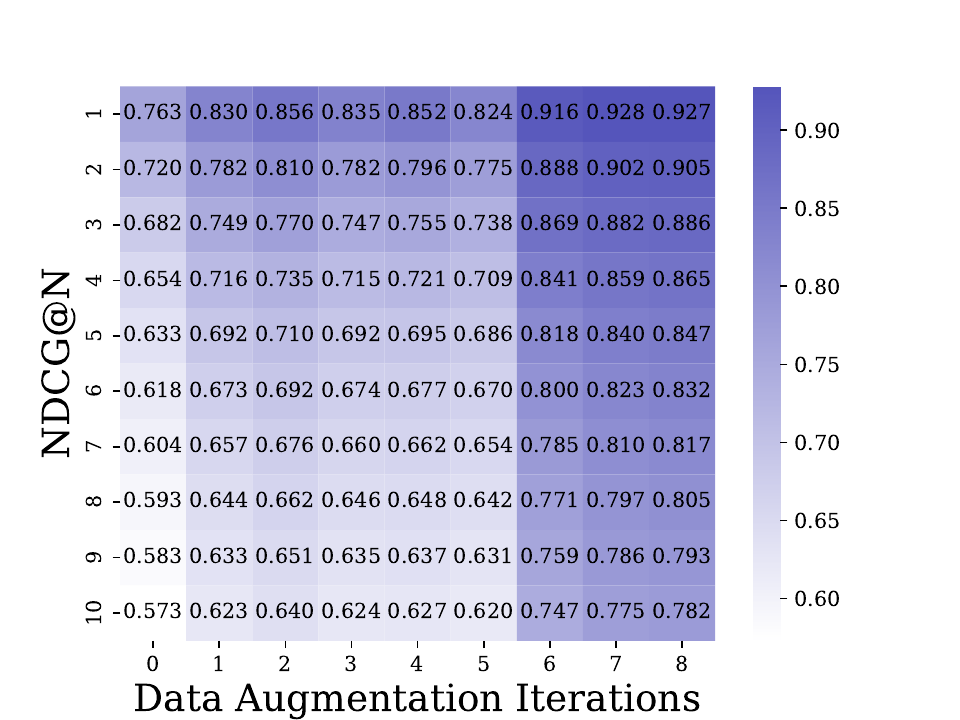}
        \caption{\ndcg}
        \label{fig:iter_ndcg}
    \end{subfigure}
    ~
    \begin{subfigure}{0.32\linewidth}
      \centering
      \includegraphics[width=\linewidth]{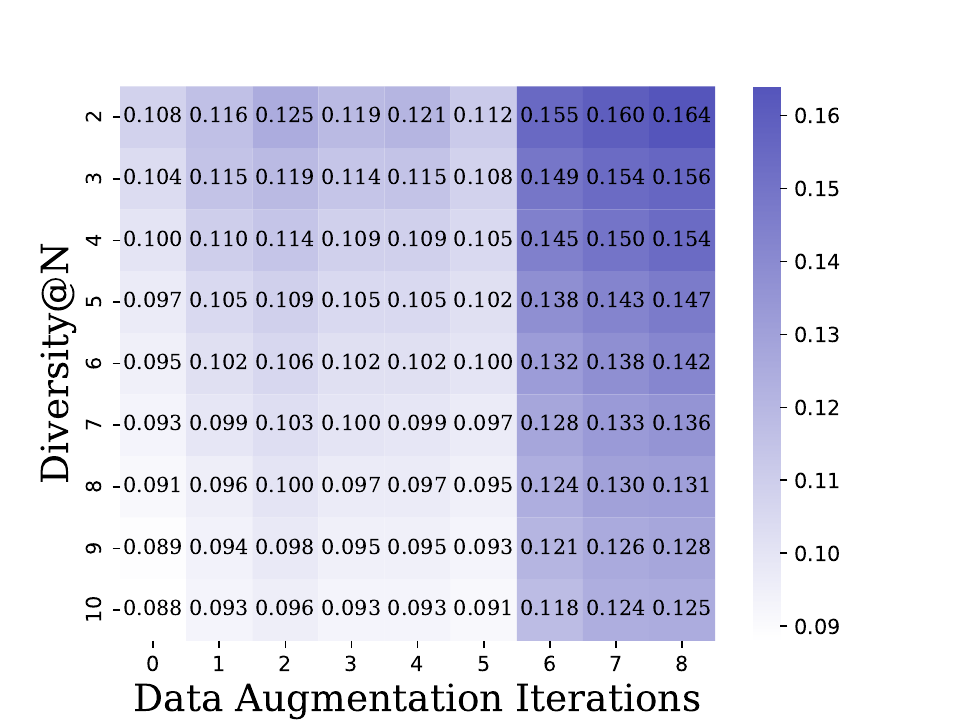}
        \caption{\diversity}
        \label{fig:iter_diversity}
    \end{subfigure}
    \caption{\method performance from data augmentation iterations; \textbf{a,}~\map; \textbf{b,}~\diversity.}
\end{figure}

Fig.~\ref{fig:iter_map} and Fig.~\ref{fig:iter_ndcg} present the performance in terms of 
\map
and \ndcg
from \method over different online iterations of data augmentation, respectively.
Note that online data augmentation is determined by 
the continuity of performance improvement on the validation set (line 16 in Algorithm~\ref{alg:augment}), 
and therefore, the performance on the test set may not be strictly improving over iterations. 
Even though, as Fig.~\ref{fig:iter_map} shows, with 8 online iterations of data augmentation, 
\method is able to improve its \map performance over all $N$ values. For example, 
\method improves
its MAP@1 performance from 0.763 at iteration 0 (i.e., the initial model using only known and random reactions) to 0.927 after 8 iterations, that is, 21.5\% improvement. 
Note that at iteration 6, the online data augmentation is switched from using only 1 new episode (line 10 in Algorithm~\ref{alg:augment}) to the {top-$5$} new episodes for each product, which
significantly boosts performance.
Similar trends are observed for \ndcg.

\begin{figure*}
    \centering
    \begin{subfigure}[t]{.32\linewidth}
      \centering
      \includegraphics[width=\linewidth]{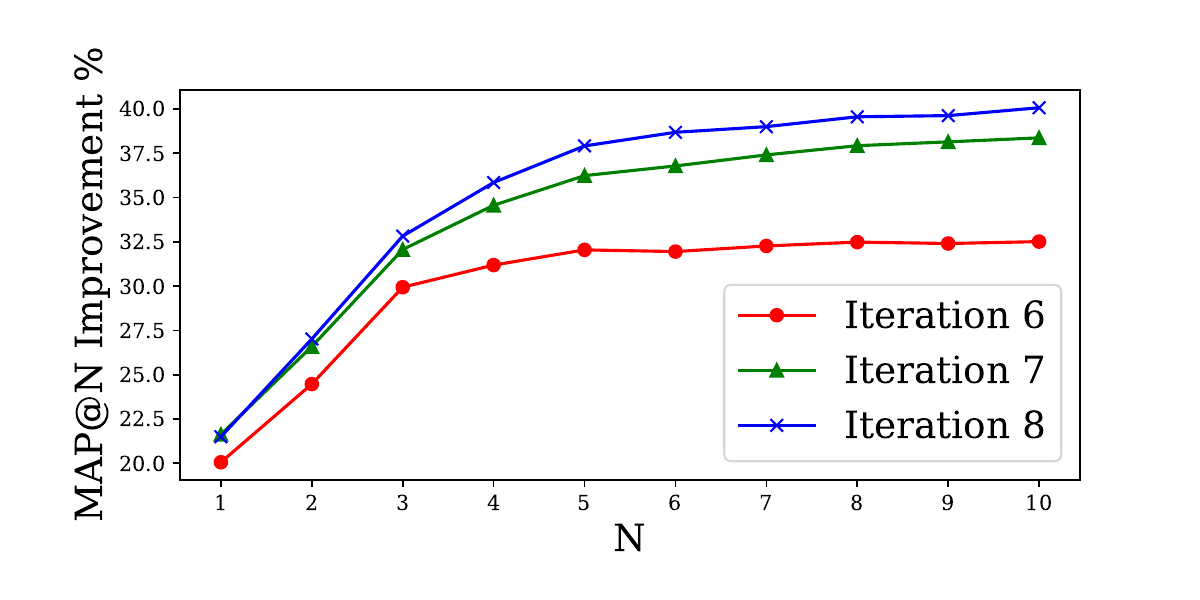}
      \caption{\map}
        \label{fig:top5_map}
    \end{subfigure}%
    ~
    \begin{subfigure}[t]{.32\linewidth}
      \centering
      \includegraphics[width=\linewidth]{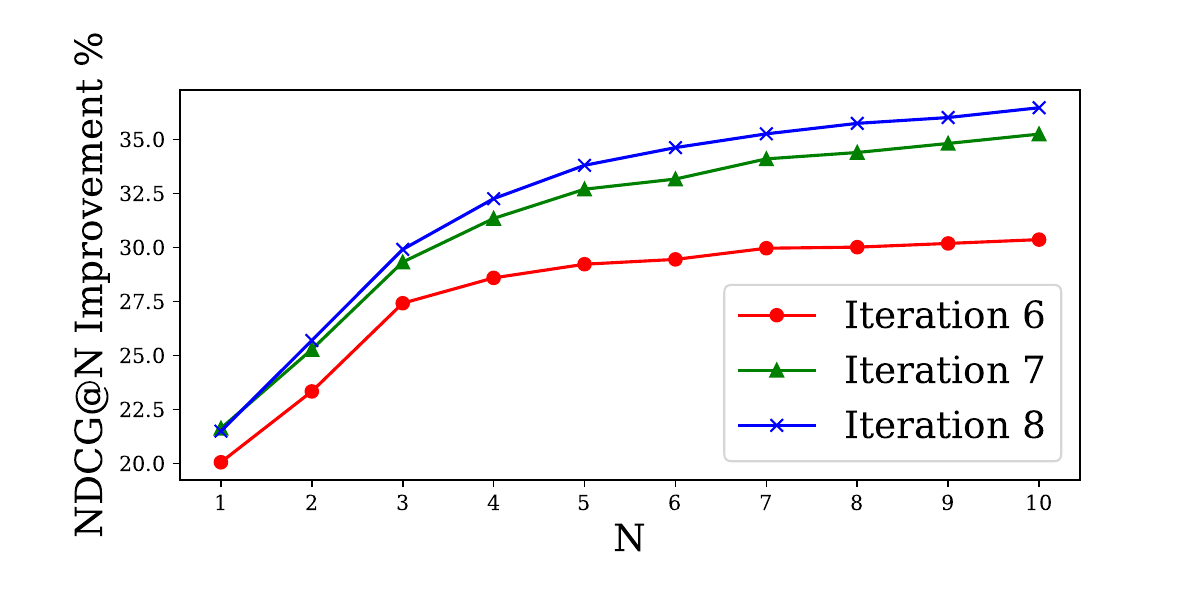}
        \caption{\ndcg}
        \label{fig:top5_ndcg}
    \end{subfigure}
    ~
    \begin{subfigure}[t]{.32\linewidth}
      \centering
      \includegraphics[width=\linewidth]{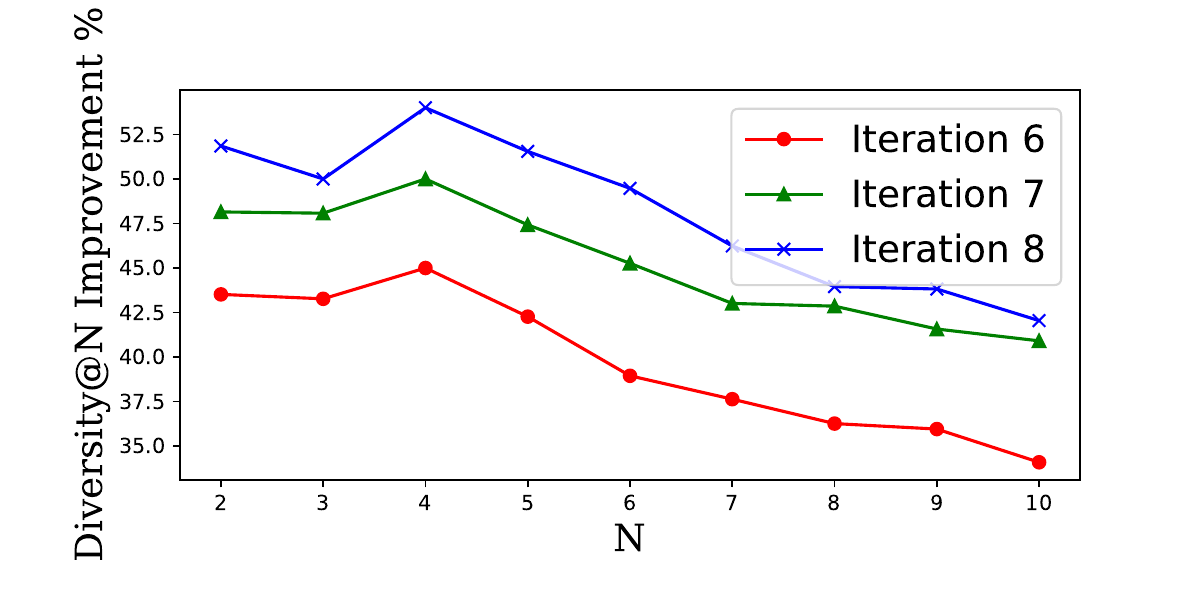}
        \caption{\diversity}
        \label{fig:top5_diversity}
    \end{subfigure}

    \caption{Improvement from Top-5 prediction search; \textbf{a,}~\map; \textbf{b,}~\ndcg; \textbf{c,}~\diversity.}
    \label{fig:top5_improv}
\end{figure*}

Fig.~\ref{fig:top5_map} and Fig.~\ref{fig:top5_ndcg} present the 
performance improvement in terms of \map and \ndcg
from iterations 6, 7 and 8 over %
the initial model, respectively.
From these figures, we see higher improvement from later iterations over the initial model. 
For example, in Fig.~\ref{fig:top5_map}, the \mbox{8-th} iteration can improve MAP@10 at 40.1\%, 
whereas the \mbox{6-th} iteration can improve MAP@10 at 32.5\%. 
Similar trends can be observed for \ndcg.
Note that 
the episodes generated for data augmentation in later iterations
are derived from agents that have been trained on data from previous iterations.
These episodes are more likely to contain correct reactions,
and subsequent agents will benefit from training on these correct reactions,
leading to better performance in later iterations.

\subsubsection{Diversity Evaluation}

Fig.~\ref{fig:iter_diversity} and
Fig.~\ref{fig:top5_diversity} present the \diversity for \method
across different data augmentation iterations.
The best diversities result from iteration 8, with Diversity@10 as {0.125} as an example.
Iterations 6, 7, and 8 provide large improvements to diversity.
These iterations augment the dataset with 5 unique episodes per product,
chosen by agents that have been trained for multiple iterations. 
In contrast, the first five iterations,
which augment the dataset with only one episode per product, do not always improve \diversity but improves
\map.

\subsection{Case Study}

\method can predict multiple reactions from the same reaction centers.
This can provide chemists with multiple options during synthetic planning.
In order to illustrate this, we highlight an example product (Fig.~\ref{fig:case:product}) 
from our test set,
the ground-truth reactants (Fig.~\ref{fig:case:gt}),
and the corresponding top-8 predictions from \method (Fig.~\ref{fig:case:top1}-\ref{fig:case:top8}).
This example is an amide coupling reaction with a diamine and a carboxylic acid as the ground truth reactants (Fig.~\ref{fig:case:gt}).
\method correctly predicts \mbox{N,N-dimethylethane-1,2-diamine} and aryl carboxylic acid
as the reactants in its top-2 prediction (Fig.~\ref{fig:case:top2}).
This prediction matches the ground-truth reactants, but is not the only synthetically useful prediction.
\method also predicts \mbox{N,N-dimethylethane-1,2-diamine}/acyl chloride (Fig.~\ref{fig:case:top1})
and \mbox{N,N-dimethylethane-1,2-diamine}/acyl bromide (Fig.~\ref{fig:case:top3}) as reactant pairs.
These are also practically useful amide coupling reactions for this product.
Interestingly, \method predicts another common reaction type in its top-7 and top-8 predictions (Fig.~\ref{fig:case:top7} and ~\ref{fig:case:top8}).
Here, \method predicts amidation reactions of esters with \mbox{N,N-dimethylethane-1,2-diamine}.
These are also synthetically useful results, although high temperatures are typically required for this reaction to occur.

However, the top-4, top-5, and top-6 predictions (Fig.~\ref{fig:case:top4}-\ref{fig:case:top6})
cannot be considered as synthetically useful because of the unusual groups present
on the acyl donor, such as the boron derivatives in the Fig.~\ref{fig:case:top6}.
Note that the top-4 and top-6 predictions still receive a reward of 1,
indicating that the forward synthesis model may reward predictions which lack practical utility.
Despite this, \method is able to provide multiple synthetically useful predictions of different reaction types from the same reaction center.
This makes \method a desirable tool for synthetic planning.

\begin{figure*}
	\centering
	\begin{subfigure}[t]{0.30\textwidth}
		\centering
		\caption{product}
		\includegraphics[width=0.95\textwidth]{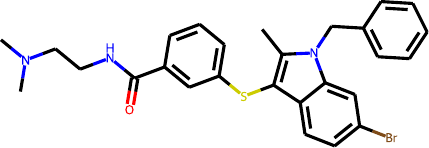}		
		\label{fig:case:product}
	\end{subfigure}
	~
	\begin{subfigure}[t]{0.30\textwidth}
		\centering
		\caption{ground-truth \\ reactants}
		\includegraphics[width=0.95\textwidth]{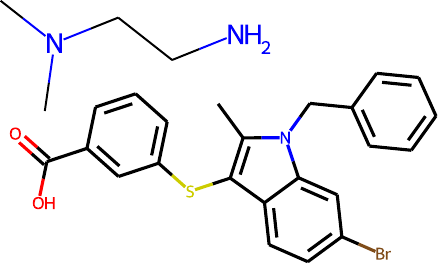}
		\label{fig:case:gt}
	\end{subfigure}
	\\
	\begin{subfigure}[t]{0.30\textwidth}
		\centering
		\caption{top-1 prediction}
		\includegraphics[width=0.95\textwidth]{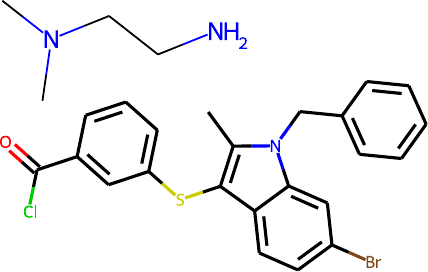}
		\label{fig:case:top1}
	\end{subfigure}
	~
	\begin{subfigure}[t]{0.30\textwidth}
		\centering
		\caption{top-2 prediction}
		\includegraphics[width=0.95\textwidth]{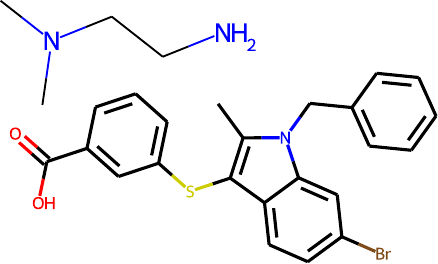}
		\label{fig:case:top2}
	\end{subfigure}
	~
	\begin{subfigure}[t]{0.30\textwidth}
		\centering
		\caption{top-3 prediction}
		\includegraphics[width=0.95\textwidth]{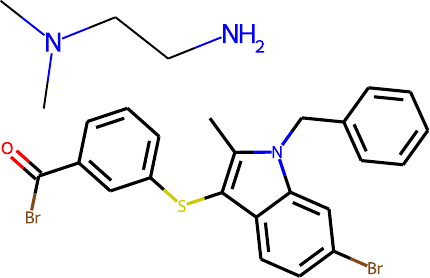}
		\label{fig:case:top3}
	\end{subfigure}
    \\
	\begin{subfigure}[t]{0.30\textwidth}
		\centering
		\caption{top-4 prediction}
		\includegraphics[width=0.95\textwidth]{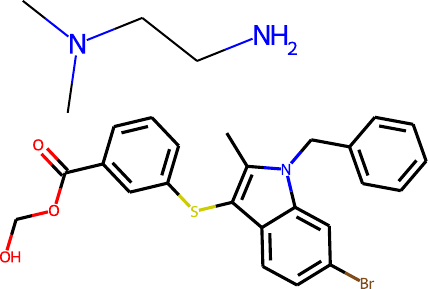}
		\label{fig:case:top4}
	\end{subfigure}
	~
	\begin{subfigure}[t]{0.30\textwidth}
		\centering
		\caption{top-5 prediction}
		\includegraphics[width=\textwidth]{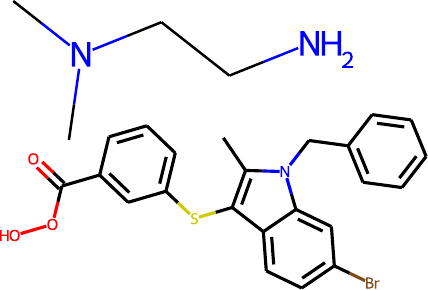}
		\label{fig:case:top5}
	\end{subfigure}
	~
	\begin{subfigure}[t]{0.30\textwidth}
		\centering
		\caption{top-6 prediction}
		\includegraphics[width=0.95\textwidth]{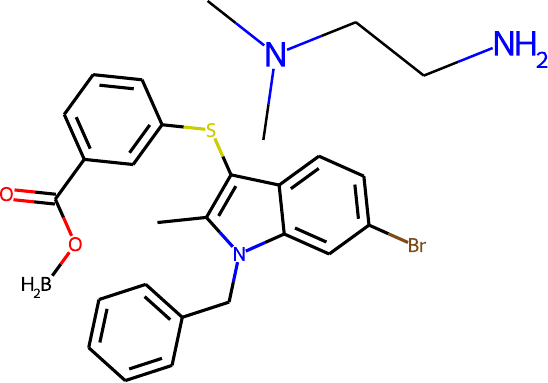}
		\label{fig:case:top6}
	\end{subfigure}
	\\
	\begin{subfigure}[t]{0.30\textwidth}
		\centering
		\caption{top-7 prediction}
		\includegraphics[width=0.95\textwidth]{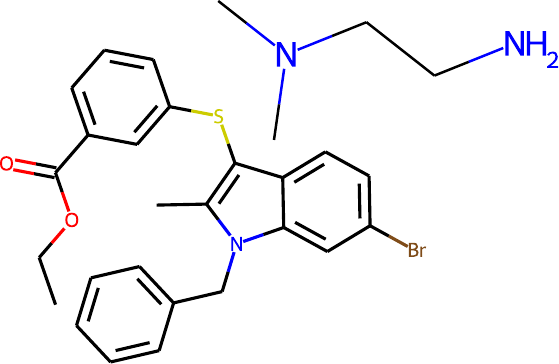}
		\label{fig:case:top7}
	\end{subfigure}
	~
	\begin{subfigure}[t]{0.30\textwidth}
		\centering
		\caption{top-8 prediction}
		\includegraphics[width=0.95\textwidth]{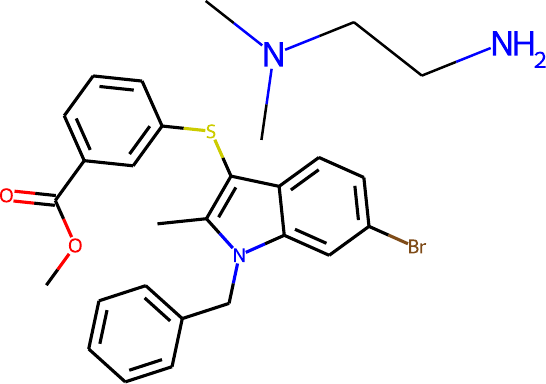}
		\label{fig:case:top8}
	\end{subfigure}
	\caption{{Predicted reactions by \method for case study;
	\textbf{a,} product; \textbf{b,} the ground-truth reactants in USPTO-50K;
	\textbf{c-j,} top predicted reactants.}}
	\label{fig:case}
\end{figure*}

\section{Discussion and Conclusions}
\label{sec:discuss}

We developed \method, a novel multi-agent reinforcement learning method with offline 
training and online data augmentation for synthon completion. 
\method has two agents to complete the two synthons of a product into reactants. 
The two agents share the same action selection policy learned from known reactions, random reactions, 
and reactions generated and deliberately selected during the online data augmentation iterations. 
Using a forward synthesis prediction model as the reward function, 
\method achieves superior performance on synthon completion
compared 
to the state-of-the-art methods. 
\method provides a new and versatile framework that can enable a more comprehensive evaluation paradigm.
Even more importantly, \method enables the exploration of reaction patterns not seen in training,
making it especially suitable for new reaction discovery purposes.

By using a reward function composed of multiple evaluation functions, 
it allows reaction evaluation with respect to the corresponding evaluation metrics, 
particularly when the reactions do not match the known reactions.  
Ideally, if high-throughput synthesis reactions can be conducted in laboratories over the predicted reactants, 
the reaction outcomes (e.g., yield) can be used as the reward.  
Even more importantly, \method enables the exploration of new reactions that are not included in the ground
truth, but are still feasible based on the reaction evaluation (i.e., the reward function), through online iterations 
of data augmentation. This feature makes \method especially suitable for new reaction discovery purposes, 
and provide reactions with respect to specific evaluation metrics.

\method uses a standalone forward synthesis prediction model in its reward function.
This reward function guides the training and exploration process,
and has a profound impact on the final model.
In our experiments, we see examples where the forward synthesis model
rewards reactions which are not practically useful to chemists.
A more precise forward synthesis model is necessary to improve our model;
however, this is outside the scope of the current work.
Fortunately, the \method framework is not dependent on a particular reward function.
When a better forward synthesis model becomes available,
\method can leverage it with minimal modification.

In retrosynthesis prediction, how to evaluate predicted reactions automatically at scale is 
under-studied~\cite{schwaller_eval_2019}. %
Existing methods compare the predictions with known reactions (i.e., ground-truth) of products, 
and consider only the predictions that exactly match the ground-truth to be correct.
However, as demonstrated in Chen {\emph{et al.}}~\cite{chen_g2retro_2023}, many ``incorrect'' predictions 
can still be chemically possible, and may even represent more viable options. 
Thus, only comparing to the ground-truth may underestimate the performance. 
Moreover, making the recovery of known reactions the only optimization objective may result in
retrosynthesis prediction models that lack the ability to discover novel reactions.

In future work, we will generalize \method for products 
with up to three reactants (i.e., up to three synthons; 
100.00\% of reactions in benchmark dataset). 
This can be done by allowing for three agents
and empty synthons if there are fewer reactants.
In these cases, the agents with empty synthons will only be able to choose \NOOP.
We will also incorporate molecular graph representation learning within \method so as to improve
its power to represent and learn from synthon and product structures. 

\section*{Data and Software Availability}

Our data and source code is publicly available online at \mbox{\url{https://github.com/ninglab/RLSynC}}.

\section*{Acknowledgements}
This project was made possible, in part, by support from the
National Science Foundation grant nos. IIS-2133650 (X.N.),
and the National Library of Medicine grant nos. 1R01LM014385-01 (X.N.).
Any opinions, findings and conclusions
or recommendations expressed in this paper are those of the
authors and do not necessarily reflect the views of the funding
agency.

\bibliography{refs}

\clearpage

\appendix
\section{Implementation Details}
\label{appendix:implementation}

We implement \method in Python 3.8.13 using PyTorch 1.12.1, RDKit 2021.03.5, and gymnasium 0.27.0.
For computing $Q$, we use a discount factor $\gamma = 0.95$ after considering $\{0.90, 0.95, 0.99\}$.
In estimating $Q$, we use a four-layer feed-forward neural network with 4,096, 2,048, 1,024, and 1 output nodes on each respective layer.
The input to the network $\statemb_{i, t}$, used 2,048-bit Morgan fingerprints of radius 2 without chirality.
We used ReLU and 0.7 dropout between each layer,
 with no activation or dropout on the output.
We explored dropout values of $\{0.1, 0.2, 0.3, 0.4, 0.5, 0.6, 0.7, 0.8\}$.
We used the Adam optimizer with learning rate $10^{-4}$ after considering $\{10^{-3}, 10^{-4}, 10^{-5}\}.$
to optimize the loss function in Equation~\ref{eqn:loss}
with a regularization coefficient of $\alpha=10^{-5}$ after considering $\{10^{-4}, 10^{-5}\}$. %
During training, we iterate over batches of $B$ products, where a training batch consists of $B\times T \times 2$ state-action pairs.
For iterations 0 (initial model), 1, 2, 3, and 4 (data augmentation), we used $B=10$.
At iteration 5, as validation performance stabilized,
we were able to see a small improvement by increasing $B$ to $20$.
For iterations 5, 6, 7, and 8, we used $B=20$.

For iterations 1, 2, 3, 4, and 5 for online data augmentation, 
we used one predicted reaction for each unique product in the training data to augment training data. 
For iterations 6, 7, and 8, we used top-5 predicted reactions ($N=5$, $k=3$ in Algorithm~\ref{alg:search}). 
When evaluating \method, we produced top-10 predicted reactions for each product in the test set
using our top-$N$ search algorithm (Algorithm~\ref{alg:search}) with $N=10$ and $k=3$.

\end{document}